# An Update Semantics for Defeasible Obligations

**Leendert van der Torre**
Department of Artificial Intelligence
Vrije Universiteit Amsterdam
De Boelelaan 1081a
1081 HV Amsterdam
The Netherlands
torre@cs.vu.nl

**Yao-Hua Tan**
EURIDIS
Erasmus University Rotterdam
P.O. Box 1738
3000 DR Rotterdam
The Netherlands
ytan@fac.fbk.eur.nl

## Abstract

The deontic logic DUS is a Deontic Update Semantics for prescriptive obligations based on the update semantics of Veltman. In DUS the definition of logical validity of obligations is not based on static truth values but on dynamic action transitions. In this paper prescriptive defeasible obligations are formalized in update semantics and the diagnostic problem of defeasible deontic logic is discussed. Assume a defeasible obligation 'normally $\alpha$ ought to be (done)' together with the fact '$\neg \alpha$ is (done).' Is this an exception of the normality claim, or is it a violation of the obligation? In this paper we formalize the heuristic principle that it is a violation, unless there is a more specific overriding obligation. The underlying motivation from legal reasoning is that criminals should have as little opportunities as possible to excuse themselves by claiming that their behavior was exceptional rather than criminal.

## 1 THE LOGIC OF NORMS

Computer scientists use the logic of obligations, prohibitions and permissions – called deontic logic – since the early eighties to represent and reason with legal knowledge (McCarty, 1994), and recently it has been used to specify and analyze security issues about electronic networks (Conte and Falcone, 1997), to represent norms in qualitative decision theory (Pearl, 1993; Boutilier, 1994; Lang, 1996) and to represent rights, duties and commitments in multi-agent systems (van der Torre and Tan, 1999b). A further increase may be expected now recently developed prescriptive deontic logics (Makinson, 1999; van der Torre and Tan, 1998a) have delivered some promising approaches for the following long-standing problems in normative reasoning and their notorious deontic paradoxes.

**Contrary-to-duty.** The conceptual issue of the contrary-to-duty paradoxes is how to proceed once a norm has been violated. Clearly this issue is of great practical relevance, because in most applications norms are violated frequently. In electronic contracting the contracting parties usually do not want to consider a violation as a breach of contract, but simply as a disruption in the execution of the contract that has to be repaired.

**Dilemma.** The conceptual problem of the dilemma paradoxes is to determine the coherence conditions of a normative system. For example, when drafting regulations a coherence check indicates whether they have this desired property, or whether they should be further modified.

In this paper we introduce a deontic update semantics for *defeasible* obligations and we show that the dynamic approach not only gives a better analysis of the traditional deontic problems, but it also gives a better analysis of the problems discussed in defeasible deontic logic. An example of reasoning with defeasible obligations is that normally you have an obligation not to have a fence around your cottage, but this obligation is defeated in the exceptional circumstances when your cottage is next to a cliff (the examples in this paper are taken from the cottage housing regulations discussed in (Prakken and Sergot, 1996)). Defeasible obligations can be overridden or cancelled by other, stronger obligations. It has been argued that more specific defeasible obligations are stronger than more general defeasible obligations, and therefore override them in case of conflict (Horty, 1993; van der Torre, 1994; Asher and Bonevac, 1996; Morreau, 1996). Unfortunately, the analysis of the specificity principle in logics of defeasible reasoning does not apply to defeasible deontic logic, because it may interfere with the violability of norms. In other words, the *combination* of reasoning about uncertainty and contrary-to-duty reasoning leads to new complications. This interference is illustrated by the following diagnostic problem.

1. There should not be a fence around the cottage.

2. If the cottage is next to a cliff, then there should be a white fence.

3. If there is a fence, then it should be white.

4. There is a white fence.

Is the fact 'there is a white fence' a violation or an exception? Obviously, this is a crucial question for legal knowledge-based systems. If the cottage is next to a cliff, then there should be a white fence according to the second line and the first obligation is cancelled. Moreover, if there is a fence, then there should be a white fence according to the third line, but the first obligation is not cancelled.

In this paper we formalize the heuristic principle that a defeasible obligation 'normally $\alpha$ ought to be (done)' together with the fact '$\neg \alpha$ is (done)' is a violation, unless there is a more specific overriding obligation. The underlying motivation from legal reasoning is that criminals should have as little opportunities as possible to excuse themselves by claiming that their behavior was exceptional rather than criminal. In absence of a cliff you have to pay a penalty for having a fence, because in that case the first obligation is a violated actual obligation. The difference between the antecedent of the second and third obligation is represented in the deontic states of the update semantics by two different orderings: the second gives rise to levels of exceptionality (inspired by preference-based approaches to defeasible reasoning) and the third gives rise to levels of ideality (inspired by preference-based approaches to deontic reasoning). Summarizing, if there is a fence without a cliff, then the first obligation is overshadowed (by the third obligation) but not cancelled (by the second obligation). It is still in force – thus it is violated.

The defeasible obligations discussed in this paper should not be confused with prima facie obligations (Ross, 1930; Asher and Bonevac, 1996; Morreau, 1996; van der Torre and Tan, 1998b). The typical example of prima facie obligations is that you have a prima facie obligation to keep your promises, but this prima facie obligation does not turn into an actual obligation when it leads to a disaster. The distinctive property is that the obligation 'there ought not to be a fence' is completely cancelled if your cottage is next to a cliff. However, if you have to break a promise to prevent a disaster, then the obligation to hold promises still holds as a prima facie obligation. Consequently, prima facie obligations have properties defeasible obligations considered in this paper do not have, such as reinstatement (van der Torre and Tan, 1997).

The layout of this paper is as follows. First, we introduce prescriptive defeasible obligations in update semantics. Second, we show that the logic formalizes the specificity principle without introducing an irrelevance problem. Third, we show how test operators can be introduced in the logic. Due to space limitations we do not discuss permissions, a first-order base language, nested conditionals, background knowledge, authorities, agents, actions, and time. In the context of deontic update semantics some of these have been discussed in (van der Torre and Tan, 1999b).

## 2   DEFEASIBLE OBLIGATIONS IN DUS

In this section we define prescriptive defeasible obligations in update semantics. The logic handles conflicts of hierarchic obligations, which normally exist, but might be dynamically re-evaluated. Two characteristic properties of the logic are that obligations are overridden by more specific and conflicting obligations, and that unresolvable strong conflicts like '$p$ ought to be (done) and $\neg p$ ought to be (done)' are 'inconsistent' in the sense that they derive all sentences of the deontic language.

We start with the basic definitions of Veltman's update semantics (Veltman, 1996). To define a deontic update semantics for a deontic language $L$, one has to specify a set $\Sigma$ of relevant deontic states (called information states in (Veltman, 1996)), and a function [ ] that assigns to each sentence $\phi$ an operation $[\phi]$ on $\Sigma$. If $\sigma$ is a state and $\phi$ a sentence, then we write '$\sigma[\phi]$' to denote the result of updating $\sigma$ with $\phi$. We can write '$\sigma[\phi_1]\ldots[\phi_n]$' for the result of updating $\sigma$ with the sequence of sentences $\phi_1, \ldots, \phi_n$. Moreover, one of the deontic states has to be labeled as the minimal deontic state, written as **0**, and another one as the absurd state, written as **1**.

**Definition 1 (Deontic update system)** *A deontic update system is a triple $\langle L, \Sigma, [\ ] \rangle$ consisting of a logical language $L$, a set of relevant deontic states $\Sigma$ and a function [ ] that assigns to each sentence $\phi$ of $L$ an operation. $\Sigma$ contains the elements **0** and **1**.*

Veltman explains what kind of semantic phenomena may successfully be analyzed in update semantics and he gives a detailed analysis of one such phenomenon: default reasoning. To define obligations in update semantics we have to define the deontic language, the deontic states and the deontic updates. The deontic language is a propositional language with the dyadic operator $\mathsf{oblige}(\alpha \mid \beta)$, read as 'normally $\alpha$ ought to be (done), if $\beta$ is (done).'

**Definition 2 (Deontic language)** *Let $A$ be a set of atoms and $L_0^A$ a propositional language with $A$ as its non-logical symbols. A string of symbols $\phi$ is a sentence of $L_1^A$ if and only if either $\phi$ is a sentence of $L_0^A$ or there are two sentences $\psi_1$ and $\psi_2$ of $L_0^A$ such that $\phi = \mathsf{oblige}(\psi_1 \mid \psi_2)$. We write $\mathsf{oblige}\ \psi$ for $\mathsf{oblige}(\psi \mid \top)$, where $\top$ stands for any tautology.*

A deontic state is a possible worlds model written as $\sigma =$

$\langle W, \leq_I, \leq_N, V \rangle$, where $W$ is a set of worlds, $\leq_I$ is an accessibility relation for ideality, $\leq_N$ is an accessibility relation for normality, and $V$ a valuation function for propositions at the worlds. For propositional $\phi$ and world $w \in W$ we write $\sigma, w \models \phi$ if the classical interpretation represented by $V(w)$ satisfies $\phi$. We add the following features to these deontic states.

**Explicit sub-state.** We extend the possible worlds model with a second deontic state, which is a sub-state of the first one. The complete state is used for the context of justification and the sub-state is used for the context of deliberation, see (van der Torre and Tan, 1998a). Whereas in Kripke semantics a unique world is singled out, called the actual world, we single out a set of worlds, called the context of deliberation.

**Full models.** We define an update system for a specific $A$, $W$ and $V$. In this paper, we assume that the deontic state contains a world for each interpretation of $L_0^A$. If we want to represent background knowledge, then this assumption has to be dropped (van der Torre, 1994; Lang, 1996).

**Non-transitive ideality relation.** We assume that the binary ideality relation is reflexive, but we do *not* assume that it is transitive or total. There is a technical problem related to the formalization of conditional obligations, discussed in (van der Torre and Tan, 1998a). A consequence of this problem is that we cannot have transitivity for the relations in the deontic states. We take the transitive closure of this relation only when we determine the preferred worlds. In (van der Torre and Tan, 1998a) we showed that a deontic state can be interpreted as a set of orderings, one for each factual sentence, instead of a unique ordering. However, this technical problem is not relevant for the intuitions of our deontic update system and the interpretation of most examples, and the deontic state can usually be identified with a single transitive ordering.

**Definition 3 (Deontic state)** *Let $L_1^A$ be a deontic language. Assume a set of worlds $W$ and a valuation function $V$ for $L_0^A$ such that for every interpretation of $L_0^A$ there is at least one corresponding $w \in W$. A deontic state is a tuple $\sigma = \langle W, W^*, \leq_I, \leq_N, V \rangle$ consisting of the set of worlds $W$, a possibly empty subset $W^* \subseteq W$, a reflexive binary relation $\leq_I$ on $W$ representing ideality, a transitive, reflexive and totally connected binary relation $\leq_N$ on $W$ representing normality, and the valuation function $V$.*

**0**, *the* minimal *state, is* $\langle W, W, W \times W, W \times W, V \rangle$,

**1**, *the* absurd *state, is* $\langle W, \emptyset, W \times W, W \times W, V \rangle$.

The deontic updates are operations on the deontic states that either zoom in on the deontic state (for facts), or create ideality and normality levels (for obligations). The prescriptive obligations have the dynamic component of creating a new deontic state. We first define the reduction of the ideality relation by an obligation. The following two definitions are extensions of definitions of the non-defeasible case in (van der Torre and Tan, 1998a). In that case, to evaluate '$\alpha$ ought to be (done) if $\beta$ is (done)' the $\alpha \wedge \beta$ and $\neg \alpha \wedge \beta$ worlds are compared. In a defeasible deontic logic, to evaluate the obligation 'normally, $\alpha$ ought to be (done) if $\beta$ is (done),' only *the most normal $\alpha \wedge \beta$* and *the most normal $\neg \alpha \wedge \beta$* worlds are compared in the ideality relation. To facilitate the definitions we assume that the normality ordering has minimal elements; the generalization to infinite descending chains is standard and straight forward (see e.g. (van der Torre and Tan, 1997)).

**Definition 4 (Reduction)** *Let $\sigma = \langle W, W^*, \leq_I, \leq_N, V \rangle$ be a deontic state, and let $W_1$ and $W_2$ be the set of respectively the most normal $\neg \alpha \wedge \beta$ and $\alpha \wedge \beta$ worlds of $W$. The reduction of $\sigma$ by oblige($\alpha|\beta$), denoted by the symbol $-$, is defined as follows.*

$$\leq'_I = \leq_I - \{w_1 \leq w_2 \mid w_1 \in W_1 \text{ and } w_2 \in W_2\}$$

$$\sigma - \text{oblige}(\alpha|\beta) = \langle W, W^*, \leq'_I, \leq_N, V \rangle$$

In the non-defeasible logic, the update $\sigma[\text{oblige}(\alpha \mid \beta)]$ is the reduction of $\sigma$ by oblige($\alpha \mid \beta$) if afterwards the best $\beta$ worlds are $\alpha$ worlds. Otherwise there is a conflict. For this definition of deontic updates we need a test whether the best (or preferred, or minimal) $\beta$ worlds are $\alpha$ worlds. This test is analogous to the satisfaction test of a dyadic obligation in the Hansson-Lewis semantics (Hansson, 1971; Lewis, 1974), and to the test whether a set of formulas preferentially entails a conclusion in preferential entailment (Shoham, 1988). In the defeasible deontic logic, we test whether the best *most normal* $\alpha \wedge \beta$ worlds are better than the best *most normal* $\neg \alpha \wedge \beta$ worlds. Note that this test is different from 'the best of the most normal $\beta$ worlds are $\alpha$ worlds' (Makinson, 1993) or 'the most normal of the best worlds are $\alpha$ worlds' which have the counterintuitive property 'what normally is (done) ought to be (done).' Definition 5 below combines the standard definition with taking the transitive closure, see (van der Torre and Tan, 1998a).

**Definition 5 (pref)** *Let $\sigma = \langle W, W^*, \leq_I, \leq_N, V \rangle$ be a deontic state, let $W_1$ and $W_2$ be the set of the most normal $\neg \alpha \wedge \beta$ and $\alpha \wedge \beta$ worlds of $W$, and let $\leq_\beta$ be the transitive closure of $\leq_I$ in $W_1 \cup W_2$, i.e. the smallest superset of $\leq_I$ such that for all $\beta$-worlds $w_1, w_2, w_3 \in W_1 \cup W_2$ with $w_1 \leq_\beta w_2$ and $w_2 \leq_\beta w_3$ we have $w_1 \leq_\beta w_3$. We write $\text{pref}(\sigma, \beta) = \alpha$ if and only if for all worlds $w_1 \in W_1$ there is a world $w_2 \in W_2$ such that $w_2 \leq_\beta w_1$ and there is no $w_3 \in W_1$ such that $w_3 \leq_\beta w_2$.*

If there is a conflict, then we introduce another exceptionality level. In this exceptional level, the previous distinction in the ideality relation is repaired.

**Definition 6 (Exception)** *Let $\sigma = \langle W, W^*, \leq_I, \leq_N, V \rangle$ be a deontic state and let $W_1$ and $W_2$ stand for the most normal $\beta$ and $\neg\beta$ worlds, respectively, and let $W_{1,1}$ and $W_{1,2}$ stand for the partition of $W_1$ in $\alpha \wedge \beta$ and $\neg\alpha \wedge \beta$ worlds. The introduction of an exceptionality level in $\sigma$ by* oblige$(\alpha|\beta)$, *denoted by the symbol $-_N$, is defined as follows.*

$\leq'_I = \leq_I + \{w_1 \leq w_2 \mid w_1 \in W_{1,1}$ and $w_2 \in W_{1,2}\}$
$\leq'_N = \leq_N - \{w_1 \leq w_2 \mid w_1 \in W_1$ and $w_2 \in W_2\}$
$\sigma -_N$ oblige$(\alpha|\beta) = \langle W, W^*, \leq'_I, \leq'_N, V \rangle$

Putting it all together we define the updates. The general principle is that in case of conflict when updating with oblige$(\alpha|\beta)$ a new exceptionality level for $\beta$ is introduced. Finally, Von Wright's contingency principle, i.e. the obligation '$\alpha$ ought to be (done) if $\beta$ is (done)' implies the consistency of $\alpha \wedge \beta$ and $\neg\alpha \wedge \beta$, is formalized by a test on the existence of $\alpha \wedge \beta$ and $\neg\alpha \wedge \beta$ worlds.

**Definition 7 (Updates)** *Let $\sigma = \langle W, W^*, \leq_I, \leq_N, V \rangle$ be a deontic state. The update function $\sigma[\phi]$ is defined as follows.*

- *if $\phi$ is a factual sentence of $L_0^A$, then*
  - *if $W' = \{w \in W^* \mid \sigma, w \models \phi\} \neq \emptyset$, then $\sigma[\phi] = \langle W, W', \leq_I, \leq_N, V \rangle$;*
  - *otherwise, $\sigma[\phi] = 1$.*

- *if $\phi =$ oblige$(\alpha|\beta)$, then*
  - *if $\sigma \neq 1$ and there are $w_1, w_2 \in W$ such that $\sigma, w_1 \models \neg\alpha \wedge \beta$ and $\sigma, w_2 \models \alpha \wedge \beta$, then*
    * *if pref$(\sigma -$ oblige$(\alpha|\beta), \beta) = \alpha$ then $\sigma[\phi] = \sigma -$ oblige$(\alpha|\beta)$*
    * *otherwise*
      · *if    pref$(\sigma -_N$ oblige$(\alpha|\beta) -$ oblige$(\alpha|\beta), \beta) = \alpha$   then   $\sigma[\phi] = \sigma -_N$ oblige$(\alpha|\beta) -$ oblige$(\alpha|\beta)$*
      · *otherwise, $\sigma[\phi] = 1$*
  - *otherwise, $\sigma[\phi] = 1$.*

A crucial notion of update systems is acceptance. The formula $\phi$ is accepted in a deontic state $\sigma$, written as $\sigma \Vdash \phi$, if the update by $\phi$ results in the same state. In that case, the information conveyed by $\phi$ is already subsumed by $\sigma$. Acceptance is the counterpart of satisfaction in standard semantics.

**Definition 8 (Acceptance)** *Let $\sigma$ be a deontic state and $\phi$ a formula of the deontic language $L_1^A$. $\sigma \Vdash \phi$ if and only if $\sigma[\phi] = \sigma$.*

If an update is accepted, then the deontic state usually has a specific content. For example, it is easily checked that a fact $\alpha$ is accepted if all the worlds of $W^* \neq \emptyset$ satisfy $\alpha$, or $\sigma = 1$.

The notion of acceptance is used to define notions of validity. An argument is $\Vdash_1$ valid if updating the minimal state **0** with the premises $\psi_1, \ldots, \psi_n$, in that order, yields a deontic state in which the conclusion is accepted, and an argument is $\Vvdash$ valid if all deontic states constructed by updating the minimal state **0** with the premises $\psi_1, \ldots, \psi_n$ *in some order such that the premises are accepted*, also accept the conclusion (van der Torre and Tan, 1998b). Note that the order of the premises is only relevant for $\Vdash_1$, not for $\Vvdash$.

**Definition 9 (Validity)** *Let $\psi_1, \ldots, \psi_n$ and $\phi$ be sentences of the deontic language $L_1^A$. The argument of $\phi$ from the premises $\psi_1, \ldots, \psi_n$ is valid, written as $\psi_1, \ldots, \psi_n \Vdash_1 \phi$, if and only if $\mathbf{0}[\psi_1] \ldots [\psi_n] \Vdash \phi$. The argument of $\phi$ from the premises $\psi_1, \ldots, \psi_n$ is nonmonotonically valid, written as $\psi_1, \ldots, \psi_n \Vvdash \phi$, if and only if for all permutations $\pi$ of $1 \ldots n$ such that $\psi_{\pi(1)}, \ldots, \psi_{\pi(n)} \Vdash_1 \psi_i$ for $1 \leq i \leq n$ we have $\psi_{\pi(1)}, \ldots, \psi_{\pi(n)} \Vdash_1 \phi$.*

It is clear that checking entailment for all possible orders in which the obligations are taken into account leads to a factorial number of entailment problems. The complexity of the inference is therefor very high.

Below some simple examples of the validity relation are given that do not create exceptionality levels, see (van der Torre and Tan, 1998a).

oblige$(p|\top) \Vvdash$ oblige$(p|q)$
oblige$(p|r),$ oblige$(q|r) \Vvdash$ oblige$(p \wedge q|r)$
oblige$(p|q \wedge r),$ oblige$(q|r) \Vvdash$ oblige$(p \wedge q|r)$
oblige$(p|\top) \not\Vvdash$ oblige$(p \vee q|\top)$
oblige$(p|q),$ oblige$(p|r) \not\Vvdash$ oblige$(p|q \vee r)$

In the following section we illustrate that one of the features of $\Vvdash$ is that more specific and conflicting obligations are only accepted if they are later than more general ones. Hence, more specific and conflicting obligations create exceptionality levels and override more general ones. Moreover, we also illustrate how the logic deals with the diagnostic problem.

## 3 EXAMPLE

Consider the two obligations from the cottage housing regulations oblige$(\neg f|\top)$ and oblige$(f|c)$, where $f$ stands for a fence around the cottage and c for a cliff next to the cottage. We first consider the situation in which the most specific obligation comes first. The ideality relation is represented in Figure 1. The corners are labelled with true atoms representing a world. For readability only positive atoms are represented. We only represent the ideality relation, because if the more specific obligation comes first then all worlds remain equivalent in the normality ordering.

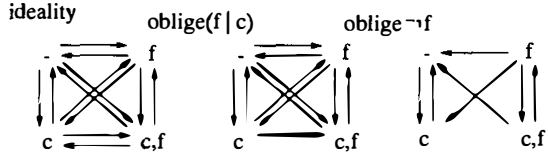

Figure 1: Fence: most specific obligation first

Otherwise, the more general obligation comes before the more specific one, as illustrated in Figure 2 below. Note that in the third ideality relation the arrows from $c \wedge \neg f$ to $c \wedge f$ are restored, because the $c$ worlds have become exceptional.

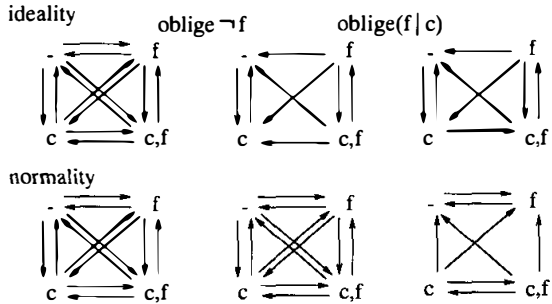

Figure 2: Fence: most general obligation first

The only accepted order of the premises in Definition 9 is that the more general obligation comes before the more specific one, because only in that case the premises are accepted.

oblige$(f|c)$, oblige$(\neg f|\top)$ $\Vdash_1$ oblige$(\neg f|\top)$

oblige$(f|c)$, oblige$(\neg f|\top)$ $\not\Vdash_1$ oblige$(f|c)$

oblige$(\neg f|\top)$, oblige$(f|c)$ $\Vdash_1$ oblige$(\neg f|\top)$

oblige$(\neg f|\top)$, oblige$(f|c)$ $\Vdash_1$ oblige$(f|c)$

The second deontic state in Figure 2 accepts the obligation oblige$(\neg f \mid c)$, but the third does not. Consequently we have the following.

oblige$(\neg f|\top)$ $\mathrel{\|\!\sim}$ oblige$(\neg f|c)$

oblige$(\neg f|\top)$, oblige$(f|c)$ $\mathrel{\|\!\not\sim}$ oblige$(\neg f|c)$

Hence, the logic formalizes the specificity principle. Moreover, the logic does not have an irrelevance problem, because we have for example that there ought to be no fence in the weekend ($w$).

oblige$(\neg f|\top)$, oblige$(f|c)$ $\mathrel{\|\!\sim}$ oblige$(\neg f|w)$

It is easily checked that the same results are obtained if we adopt the first two formulas of the example in the introduction, i.e. oblige$(\neg f|\top)$ and oblige$(w \wedge f|c)$ where $w \wedge f$ stands for a white fence. This is a consequence of the fact that we use tests on minimal worlds by testing whether pref$(\sigma, \beta) = \alpha$. The difference is the constructed ideality relation of $c$ worlds.

oblige$(\neg f|\top)$ $\mathrel{\|\!\sim}$ oblige$(\neg f|c)$

oblige$(\neg f|\top)$, oblige$(w \wedge f|c)$ $\mathrel{\|\!\not\sim}$ oblige$(\neg f|c)$

Finally, we consider the diagnostic problem by taking the third (and contrary-to-duty) oblige$(w \wedge f|f)$ into account. The only orders that accept all premises are again orders in which the second obligation is later than the first obligation, which leads to the following three orders.

oblige$(w \wedge f|f)$, oblige$(\neg f|\top)$, oblige$(f|c)$

oblige$(\neg f|\top)$, oblige$(w \wedge f|f)$, oblige$(f|c)$

oblige$(\neg f|\top)$, oblige$(f|c)$, oblige$(w \wedge f|f)$

The first two orders lead to a deontic state in which all $w \wedge f$ worlds are preferred to $\neg w \wedge f$ worlds, whereas the latter order leads to a state in which only the $\neg c \wedge w \wedge f$ worlds are preferred to $\neg c \wedge \neg w \wedge f$ worlds. The latter order is represented in Figure 3. In the figure we do not represent $f \wedge \neg w$ worlds, because they are meaningless (they could be deleted from the model by using background knowledge, see (van der Torre, 1994; Lang, 1996)). Moreover, with $(w)$ we represent that $w$ can either be true or false.

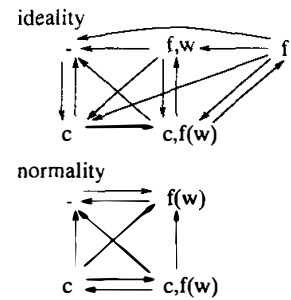

Figure 3: Fence with contrary-to-duty

Figure 3 illustrates how the deontic update semantics for defeasible obligations deals with the diagnostic problem. As explained in (van der Torre and Tan, 1998b), to deal with the diagnostic problem we have to distinguish be-tween decision variables and parameters or events (Lang, 1996) (called controllable and uncontrollable propositions in (Boutilier, 1994)). We therefor call $c$ a parameter and $w$ and $f$ decision variables. There is a violation of a duty proper if and only if for every world that still can be realized and that is most normal there is a better world which can no longer be realized and which is as least as normal, and that has the same truth values for the parameters.

First, consider the most normal $w \wedge f$ state, i.e. $\neg c \wedge w \wedge f$. Though this state is not the worst state (this is $\neg c \wedge \neg w \wedge f$), it is clear that this state is a violation state. The $\neg c \wedge \neg f$ state is preferred to it, it is as normal, and it has the same truth values for the parameter $c$. Moreover, consider $c \wedge f$. The only world that has the same values for the parameters is $c \wedge \neg f$, but these worlds are worse. Consequently $c \wedge f$ is *not* a violation.

If we replace the second obligation by oblige($w \wedge f \mid c$), then the same argument goes through. However, now the ideality order of the $c$ worlds is identical for each order of the premises below: $c \wedge w \wedge f$ is preferred to $c \wedge \neg f$, which is preferred to $c \wedge \neg w \wedge f$.

oblige($w \wedge f \mid f$), oblige($\neg f \mid \top$), oblige($w \wedge f \mid c$)

oblige($\neg f \mid \top$), oblige($w \wedge f \mid f$), oblige($w \wedge f \mid c$)

oblige($\neg f \mid \top$), oblige($w \wedge f \mid c$), oblige($w \wedge f \mid f$)

Summarizing, in the semantics different orderings are introduced to distinguish violations and exceptions. They correspond to so-called factual and overridden defeasibility (van der Torre and Tan, 1997). Several other cases of apparent dilemmas of defeasible obligations can be solved by analyzing them as different types of defeasibility. Examples of conflict defeasibility are 'be polite' and 'be honest' ({oblige $p$, oblige $h$}), and 'the window ought to be closed if it rains' and 'it ought to be open if the sun shines' ({oblige($c \mid r$), oblige($\neg c \mid s$)}). Examples of factual defeasibility are the contrary-to-duty paradoxes, such as 'Smith should not kill Jones,' 'if Smith kills Jones, then he should do it gently' and 'Smith kills Jones'

{oblige $\neg k$, oblige($g \mid k$), $k$}

given that gentle killing implies killing ($\vdash g \rightarrow k$), and 'a certain man should go to the assistance of his neighbors,' 'if the man goes to their assistance, then he should tell them that he will come,' 'if the man does not go to the assistance, then he should not tell them he will come' and 'the man does not go'

{oblige $a$, oblige($t \mid a$), oblige($\neg t \mid \neg a$), $\neg a$}

In the following section we consider the latter example in the deontic update semantics extended with test operators.

## 4   TEST OPERATORS

In this section we formalize test operators in the deontic update semantics. We write ideal($\alpha \mid \beta$) for the test 'ideally, $\alpha$ is (done), if $\beta$ is (done).' The interaction between oblige and ideal is analogous to the interaction between normally and presumably operators in Veltman's update semantics (Veltman, 1996). The prescriptive obligations oblige($\alpha \mid \beta$) have the dynamic component of creating a new deontic state, whereas the tests evaluate what the norms are in a particular deontic state. The reason to introduce this new operator is that the ideal operator has weakening of the consequent and the disjunction rule for the antecedent. In Section 2 we showed that the prima facie obligations do not have weakening of the consequent, but they do have (restricted) strengthening of the antecedent. It is shown in (Tan and van der Torre, 1996) that we have to introduce a separate operator for weakening of the consequent to combine these two intuitive inference patterns. For example, if the obligations oblige($\alpha \mid \beta$) also have weakening of the consequent, then the counterintuitive obligation oblige($\neg(w \wedge f) \mid f$) can be derived from oblige($\neg f \mid \top$) by respectively weakening of the consequent and strengthening of the antecedent, regardless of the existence of another premise oblige($w \wedge f \mid f$). The deontic language $L_1^A$ is extended with the two dyadic operators ideal($\alpha \mid \beta$) and ideal$^*(\alpha \mid \beta)$.

**Definition 10 (Deontic language, continued)** *Let $A$, $L_0^A$ and $L_1^A$ be as defined in Definition 2. A string of symbols $\phi$ is a sentence of $L_2^A$ if and only if either $\phi$ is a sentence of $L_1^A$ or there are two sentences $\psi_1, \psi_2$ of $L_0^A$ such that either $\phi = $ ideal($\psi_1 \mid \psi_2$) or $\phi = $ ideal$^*(\psi_1 \mid \psi_2)$. We write ideal $\psi$ for ideal($\psi \mid \top$) and ideal$^*$ $\psi$ for ideal$^*(\psi \mid \top)$.*

For the dynamic interpretation of a test we define it analogously to the test operator in dynamic logic, and to the might and presumably operators in Veltman's update semantics (Veltman, 1996). If the test is successful then the information conveyed by the test is already subsumed by the deontic state and the test update simply returns the state. Otherwise the test update returns the absurd state.

**Definition 11 (Deontic updates, continued)** *Let $\sigma = \langle W, W^*, \leq_I, \leq_N, V \rangle$ be a deontic state, and let pref$^*$ be defined as pref, but with the worlds restricted to $W^*$ instead of $W$. The update function $\sigma[\phi]$ defined in Definition 7 is extended as follows.*

- *if $\phi = $ ideal($\alpha \mid \beta$), then*
  - *if pref($\sigma, \beta$) $= \alpha$, then $\sigma[\phi] = \sigma$; otherwise, $\sigma[\phi] = \mathbf{1}$.*

- *if $\phi = $ ideal$^*(\alpha \mid \beta)$, then*
  - *if pref$^*(\sigma, \beta) = \alpha$, then $\sigma[\phi] = \sigma$; otherwise, $\sigma[\phi] = \mathbf{1}$.*

...The following contrary-to-duty paradox called Chisholm's paradox (Chisholm, 1963) illustrates how the two operators oblige and ideal are combined, and a fortiori how the two inference patterns strengthening of the antecedent and weakening of the consequent are combined. The contrary-to-duty paradoxes are important benchmark problems of deontic logic. The examples also illustrate that the deontic contrary-to-duty paradoxes can be modeled without any problems in DUS, and that the dynamic representation is in some respects more insightful than the static one.

Consider the sentences 'a certain man should go to the assistance of his neighbors' (oblige $a$), 'if the man goes to their assistance, then he should tell them that he will come' (oblige$(t \mid a)$), 'if the man does not go to the assistance, then he should not tell them he will come (oblige$(\neg t \mid \neg a)$)) and 'the man does not go' ($\neg a$). Regardless of the ordering of the premises, the ideality relation in Figure 4 results. There are no conflicts and therefor no exceptionality levels are introduced. All worlds are equivalent in the normality ordering.

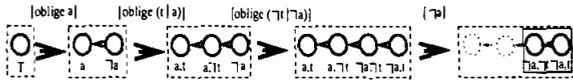

Figure 4: Chisholm's paradox

In the context of justification ($W$) we have that in the ideal state the man tells his neighbors ($t$),

oblige $a$, oblige$(t \mid a)$, oblige$(\neg t \mid \neg a)$, $\neg a$ ⊩ ideal $t$

and in the context of deliberation ($W^*$) we have that in the ideal state the man does not tell his neighbors ($\neg t$).

oblige $a$, oblige$(t \mid a)$, oblige$(\neg t \mid \neg a)$, $\neg a$ ⊩ ideal$^*$ $\neg t$

We finally show how strengthening of the antecedent and weakening of the consequent are combined. Strengthening of the antecedent is a property of oblige and weakening of the consequent is a property of ideal. We have:

oblige $a$, oblige$(t \mid a)$, oblige$(\neg t \mid \neg a)$, $\neg a$ ⊩ oblige$(a \wedge t)$
oblige $a$, oblige$(t \mid a)$, oblige$(\neg t \mid \neg a)$, $\neg a$ ⊮ oblige $t$
oblige $a$, oblige$(t \mid a)$, oblige$(\neg t \mid \neg a)$, $\neg a$ ⊩ ideal$(a \wedge t)$
oblige $a$, oblige$(t \mid a)$, oblige$(\neg t \mid \neg a)$, $\neg a$ ⊩ ideal $t$

The conclusion ideal $t$ is derived from ideal$(a \wedge t)$, which is derived from oblige$(a \wedge t)$. See (Tan and van der Torre, 1996) for a further discussion on these two phases of deontic reasoning.

## 5 CONCLUDING REMARKS

The dynamic approach to formalizing norms (van der Torre and Tan, 1998a) follows a recent trend in dynamic semantics (van Benthem, 1996). In this paper it has been extended by a logic of prescriptive defeasible obligations, in which more specific obligations override more general ones. The specificity problem is solved by making more specific obligations refer to exceptional circumstances.

Most of the deontic paradoxes concern obligations, and therefore most of the paradox driven research in deontic logic has focussed on obligations. However, in some applications – e.g. computer security – rights and permissions play a dominant role. In 'standard' approaches permission is defined as the absence of obligation. However, there are at least the following two problems with this so-called 'weak' permission.

**Free choice.** The conceptual problem of the free choice paradoxes is whether a permission for a disjunctive sentence implies the permission of each of its disjuncts (von Wright, 1968; Kamp, 7374). This property cannot be accepted for weak permissions, and therefore new free choice operators have been defined.

**Strong permission.** A so-called 'strong' permission – the existence of an explicitly permitting norm – does not follow from the absence of an obligation, because an act or state can be neither obligatory nor explicitly permitted. Thus it cannot be defined as weak permission, and again new operators have been defined.

Strong permissions have been defined in (van der Torre and Tan, 1999b). These operators can be restricted to the most normal worlds, and then they can be used in the update semantics proposed in this paper.

In the update semantics for prima facie obligations (van der Torre and Tan, 1998b) a value (its strength) is associated with prima facie obligations (Ross, 1930), such that they can override weaker obligations (van der Torre and Tan, 1997). They are represented by deontic states where the ideality relation is replaced by a ranking function on ordered pairs of worlds, i.e. by $\langle W, R, V \rangle$, where $R$ is a ranking function from $W \times W$ to the set of values. It is an open question how the two types of overridden defeasibility can be combined in deontic states that have a ranking function as well as a normality relation $\langle W, W^*, R_I, \leq_N, V \rangle$.

It is well known that defeasible deontic logic is related to logics for qualitative decision theory (Pearl, 1993; Boutilier, 1994; van der Torre and Tan, 1999a). In particular, prima facie obligations are related to logics of desires (qualitative abstractions of utilities) which can be overridden by stronger desires (Tan and Pearl, 1994; Lang, 1996). The formal relation between obligations and desires is subject of present investigations.